\documentclass[10pt,twocolumn,letterpaper]{article}

\usepackage{cvpr}
\usepackage{times}
\usepackage{epsfig}
\usepackage{graphicx}
\usepackage{amsmath}
\usepackage{amssymb}
\usepackage{tabularx}
\usepackage{enumitem}
\usepackage{caption}
\usepackage{epstopdf}
\usepackage[backend=bibtex,sorting=none,natbib=true,style=numeric]{biblatex}
\usepackage{pdfpages}
\usepackage[english]{babel}
\usepackage{listings}
\usepackage{float}

\cvprfinalcopy 

\def\cvprPaperID{****} 

\bibliography{report_short}
\begin{document}

\title{Robust Large-Scale Localization in 3D Point Clouds Revisited}
\author{Fabian Tschopp\\
	{\tt\small tschopfa@student.ethz.ch}
	\and
	Marco Zorzi\\
	{\tt\small mzorzi@student.ethz.ch}
}

\maketitle

\begin{abstract}
We tackle the problem of getting a full 6-DOF pose estimation of a query image inside a given point cloud. This technical report re-evaluates the algorithms proposed by Y. Li \etal ``Worldwide Pose Estimation using 3D Point Cloud'' \cite{li2012}. Our code computes poses from 3 or 4 points, with both known and unknown focal length. The results can easily be displayed and analyzed with Meshlab. We found both advantages and shortcomings of the methods proposed. Furthermore, additional priors and parameters for point selection, RANSAC and pose quality estimate (inlier test) are proposed and applied.
\end{abstract}

\section{Introduction}
Given an Stucture-from-Motion point cloud, this project estimates the camera pose (position and orientation) from which a new photo was taken. This is done by matching 2D-3D features between the image and the point cloud \cite{li2012,sattler2012}. The problem with previous approaches is that finding good matches is hard due to similar local structures, especially when considering large data sets. We aim at resolving this by relaxing the matching criterion, incorporating visibility of matches from camera perspectives and bidirectional matching features from 3D to 2D to find additional good matches. The input to our implementation is a Structure-from-Motion point cloud with SIFT descriptors and a set of images to process. The images already have their SIFT features extracted and stored in separate keyfiles. The outputs are camera poses for each processed picture, reprojected visualizations and a point cloud/mesh for viewing in Meshlab.
The project implements these approaches according to the paper by Y. Li \etal \cite{li2012}.

\section{Implementation}
\subsection{Basic approach}
\label{basic_approach}
The basic approach uses a simple RANSAC scheme, FLANN kd-tree and P3P/P4P. This part of the code was necessary in order to have a comparison basis for the new approach described by Y. Li \etal \cite{li2012}.
We use the FLANN library to perform nearest neighbor search of query SIFT features in the kd-tree. From this we get matches between the query image and the point cloud which need verification. Lowe's ratio test (with ratio 0.7) is used to classify this matches as ``good'' matches, meaning that they can be used for the RANSAC algorithm. The RANSAC is started with the following standard parameters:

\begin{itemize}[noitemsep]
	\item Maximum number of iterations: 10'000.
	\item $L_2$ distance (reprojection) for inliers: $0.5\,m$.
	\item Fraction of good matches as inliers (fitted matches) to stop RANSAC: $1/10^{\text{th}}$.
	\item Total amount of good matches as inliers (fitted matches) to stop RANSAC: 12.
\end{itemize}

The basic approach starts by selecting 3 or 4 (without known focal length) unique points from the set of good matches previously identified. We then use the algorithms from M. Bujnak \etal \cite{p4p} for the 4 points problem and the one from L. Kneip \etal \cite{p3p} for the three point problem. 
Those algorithms, though, give all the mathematically viable solutions which need validation. In order to do this, we start by centering the features around the image center and project the 3D features onto the image and measure the $L_2$ error of this back projection.
With the matches that fit according to the projection, we assess the quality of the pose as following:
\begin{itemize}[noitemsep]
	\item $G, F$: Sets of good and fit matches.
	\item $w, h$; Image width and height.
	\item $c=\frac{w}{40}$: Half of the area size covered by a match.
	\item $m_x, m_y$: Match x and y coordinate on the picture.
	\item $I^{g}_{i,j}$: Image map of covered areas by good matches.
	\item $I^{f}_{i,j}$: Image map of covered areas by fitted matches.
	\item $A^{g}$: Image area covered by good matches.
	\item $A^{f}$: Image area covered by fitted matches.
\end{itemize}

\begin{align}
	\forall m \in G.\,i=m_x-c,\ldots,m_x+c,\\
	j=m_y-c,\ldots,m_y+c\qquad I^g_{i,j}=1\\
	\forall m \in F.\,i=m_x-c,\ldots,m_x+c,\\
	j=m_y-c,\ldots,m_y+c\qquad I^f_{i,j}=1
\end{align}
\begin{align}
	A^{g} = \sum_{i=1,j=1}^{i=w,j=h}I^{g}_{i,j}\\
	A^{f} = \sum_{i=1,j=1}^{i=w,j=h}I^{f}_{i,j}\\
	q = \frac{A^{f}}{A^{g}}
\end{align}

This quality estimate has a lower bound of 0 and an upper bound of 1. It is used to tell if one solution is better than another. The amount of fitted matches does not matter for selecting the best solution, only the quality estimate does.
After assessing that the quality of the current solution is better than the previous one we save the solution as the most viable. The whole RANSAC process is repeated until the exit conditions apply or all steps are used up.
\subsection{Advanced approach}
\label{advanced_approach}
The advanced approach aims, as mentioned in the paper by Y. Li \etal \cite{li2012}, to create a robust and scalable algorithm for worldwide size point clouds. In order to be able to benchmark the two approaches, we focused on implementing the bidirectional matching and the co-occurrence prior sampling techniques mentioned, and tuned them with our own parameters.
We kept the same code for the construction of the kd-tree. When coding the Lowe's test, we also save the cameras in which the dataset point was observed. This is necessary for the improved RANSAC algorithm, where camera set intersections are computed. The Lowe's test ratio is now 0.9, and the co-occurence prior will discriminate wrong matches sufficiently.
We then begin the RANSAC algorithm with the following parameters:

\begin{itemize}[noitemsep]
	\item Exact number of iterations: 100 (no backmatching), 200 (with backmatching).
	\item $L_2$ distance (reprojection) for inliers: $0.5\,m$.
	\item Fraction of good matches as inliers (fitted matches) to skip backmatching: $1/10^{\text{th}}$.
	\item Total amount of good matches as inliers (fitted matches) to skip backmatching: 12.
\end{itemize}

Note that the stopping conditions for our advanced approach are different, and a set of 12 fitted matches will not stop RANSAC, but only skip the backmatching. This will lead to better quality solutions while still having short running times, as elaborated in section \ref{benchmark}.

The paper by Y. Li \etal \cite{li2012} introduces two new techniques which we used:
\begin{enumerate}[noitemsep]
	\item Co-occurrence of 3D model points in images to improve RANSAC.
	\item Bidirectional matching (backmatching) of 3D model points with image features.
\end{enumerate}

The co-occurrence prior consists of pre-filtering points in order to reduce RANSAC iterations and improve performance. We draw points sequentially and then decide whether to accept the point or not by associating a probability proportional to the size of the intersection. 

\begin{equation}
\label{EQN:proportional}
 \tilde{P}r_{select}(p_i | p_1, ..., p_{i-1}) \propto | A_{p_1} \cap ... \cap A_{p_i} |
\end{equation}

Although equation \ref{EQN:proportional} and its origin ar well explained in the paper by Y. Li \etal \cite{li2012}, nothing is said about how the probability should be calculated. We decided to compute it as follows:

\begin{align}
f_{scaling} = \frac{1}{1+e^{\frac{-| A_{p_1} \cap ... \cap A_{p_i} |}{k}}} \\
f_{ratio} = \frac{| A_{p_1} \cap ... \cap A_{p_i} |}{\min(| A_{p_1} \cap ... \cap A_{p_{i-1}} |,| A_{p_i} |)} \\
\tilde{P}r_{select}(p_i | p_1, ..., p_{i-1}) = f_{scaling} \cdot f_{ratio}
\label{EQN:probability}
\end{align}
This allows equation \ref{EQN:probability} to get 75\% acceptance probability pre-multiplier when \begin{math} | A_{p_1} \cap ... \cap A_{p_i} | = k \end{math} which we set to 5. Note that we use this prior because the second term of the equation is independent of absolute set sizes. We did however want to favor bigger intersections and still not completely exclude smaller intersections.
To avoid being stuck in dead end intersections, we force these conditions:
\begin{itemize}[noitemsep]
	\item If the intersection is zero for more than 30 times, all points get discarded and a new first one is chosen.
	\item We start with a point having at least 5 cameras.
\end{itemize}

After these tests are passed, we get the 3 or 4 points and perform the same reprojection tests as in the basic approach.

If all the RANSAC steps are used up and no solution is found, we try backmatching and reset the RANSAC steps.

The backmatching steps are the following, as found in the paper by Y. Li \etal \cite{li2010}, and tuned by us:

\begin{enumerate}[noitemsep]
\item Creation of a new FLANN kd-tree based on image features.
\item Setting the following parameters:
	\begin{itemize}[noitemsep]
	\item Number of nearest neighbors to look for: 2 
	\item Number of backmatches to be achieved: 100
	\item Lowe's ratio test value: 0.7
	\item Dynamic priorities booster: 10
	\end{itemize}
\item Determine the maximum priority of matches from the view graph to boost the good matches to the top of the queue.
\item Boosting of the good matches to the top of the queue.
\item Prioritized picking of a 3D feature point.
\item Ratio test and accept/reject the backmatch.
\item Addition of  all views to those that we have to increase in priority.
\item Update priority queue according to prioritized views. This boosts additional 3D features other than the good matches to the top of the priority queue.
\end{enumerate}
After the backmatching is finished, RANSAC reruns and the best solution with the fitted matches are stored in the query.

\subsubsection{Method advantages}
This method aims to solve the scalability issue for huge point clouds and taking it to a world size scale. It allows to robustly scale the point cloud, including multiple cities and still get good pose estimations. This is due to the good discrimination of bad matches with the co-occurence prior and additionally finding good matches through backmatching. We were able to see some of the claimed advantages in our benchmarking section \ref{benchmark}.

\subsubsection{Method shortcomings}
The advanced method cannot improve much on the Dubrovnik dataset alone. Better quality is found but there are also has a lot more parameters to tune. The algorithm proposed would be most useful when multiple datasets with similar features (such as Rome and Dubrovnik) are fused together.

However, testing this would require to fuse datasets and re-index all matches 6and cameras. This would have been possible for us, however, the increased memory usage would not have been viable on our machines and could quickly rise to over 16 GB. For a worldwide pose estimation, multiple FLANN kd-trees and distribution across clusters seems to be necessary, but not hard to achieve.

\section{Program structure}

\subsection{Program runtime}
As soon as the program starts, the golden dataset (original bundler output) and the information dataset (excluding query poses) are loaded. We use the Dubrovnik dataset for testing purposes because it is the smallest in size, thus making debugging faster. It contains about 2 million 3D points and takes about 15 seconds to load on recent personal computers. After this process, the user is required to chose between the basic approach or the advanced one.

After selection, the query processor object is created, where SIFT features get averaged per 3D point and a kd-tree using the FLANN library is initialized.
Then, we load the provided list of query images so to ask the user which query image needs to be evaluated.

The benchmark tests the algorithms quality and efficiency: It computes the poses of all the query images and measures the time. For the quality we measure the error against the golden pose estimation and compute the norm of the error in terms of rotation, translation and focal length.

Where possible, the code runs in parallel, using the OpenMP library.

\subsection{Visualization export formats}
The mesh is exported as \textit{ply} (Polygon File Format, Stanford Triangle Format). It includes all points from the point cloud of the bundler output, including color information for better visualization. It can be loaded using Meshlab.
For a camera pose, several different files are exported:
\begin{itemize}[noitemsep]
	\item \textit{camera.mlp}: Meshlab project file, specifying a virtual meshlab camera so that the situation in which the picture was taken can be simulated with the 3D point cloud. The original picture can be displayed as overlay over the virtual pose, giving a good visualization of the camera pose quality.
	\item \textit{camera.obj}: Wavefront OBJ format, containing a small camera 3D model to show the camera's pose as well as a planar sprite to display the original picture.
	\item \textit{camera\_proj.obj}: Wavefront OBJ format, containing edges that origin from the camera, peek through the image and connect to the matched points in the 3D point cloud.
	\item \textit{camera.jpg}: The original query image in original size as found on \textit{flickr.com}, but excluding metainformation.
\end{itemize}

\begin{figure}[H]
	\centering
	\includegraphics[width=0.7\linewidth]{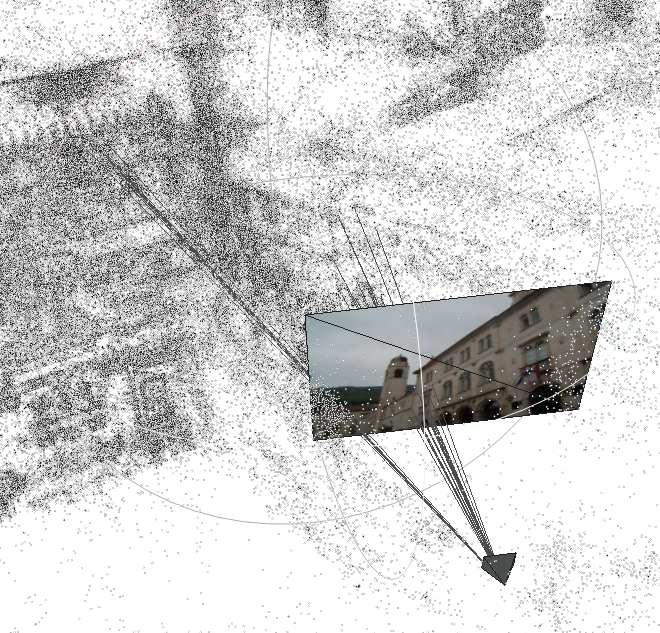}
	\caption{Meshlab visualization with camera, sprite and projection edges.}
	\label{FIG:sc001}
\end{figure}

\section {Difficulties}
\subsection{Code translation from Matlab to C++}
Since the P4pf code was given in a paper, it needed to be translated to Matlab. This led to difficulties because Matlab fits very well for matrices manipulation, while C++ improves efficiency. In order to have the same results we had to explore the Eigen library thoroughly.

\subsection{Export from dataset to visualization}
The visualization was not trivial to implement because Meshlab does not have a good documentation on its file format. This also included rotating and flipping the 3D axes until the solution matched, and inserting the right viewport and focal length. Also, we had to merge the information from the point cloud and the golden dataset since SIFT descriptors and colors of the point cloud were stored separately. As a result, we had to merge these information to get a complete output file (3D mesh).

\subsection{Camera pose reference frame}
More difficulties were found while visualizing the first results. We had correct matches but flipped pictures or even flipped poses due to the fact that there was not a coherent reference frame among all the algorithms. We firstly had to localize were this was happening and, subsequently, we had to flip some axis in order to get a correct pose and visualization. Our reference axis are so that the image is centered at zero, the camera looks at the +Z axis and +X, +Y are oriented towards the right and up side of the image. Compared to this, bundler looks at -Z. Also, P3P and P4P give different results: P3P gives camera position and translation, while P4P gives the world to camera transformations.

\subsection{ Set intersection for improved RANSAC }
As aforementioned, the set intersection proportionality was not easy to understand, also because the paper \cite{li2012} does not mention any of the problems we encountered. After analyzing all the set sizes, we found out that most of the matches have either 2, 3 or 4 cameras. Some query images have higher values but it is fair to say that there is a tendency of the 3D points to be observed in less than 5 cameras. As a result, we found out that if the set $A_{p_i}$ had a small size, it would be very unlikely to have an intersection of 3 or 4 points.

\section{Quality}
We had some issues with the quality of the advanced processor: The co-occurrence prior quickly found solutions with 12 or more fitted matches. However, the poses often were based on points that lied in a far-plane such as the mountain background of Dubrovnik, or were generally very clustered together in the picture. This meant the pose was approximately right in terms of rotation, but the pose often more than $100\,m$ wrong.

To fix this, we changed from using the fitted matches set size as quality estimate to an image coverage quality estimate as mentioned in section \ref{basic_approach}.

\begin{figure}[htb]
	\centering
	\includegraphics[width=0.7\linewidth]{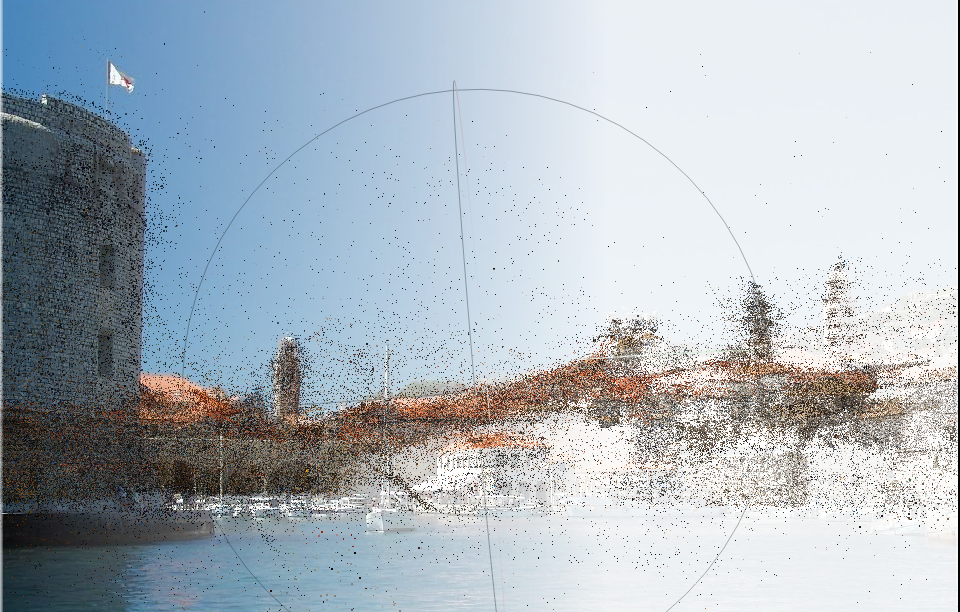}
	\caption{Meshlab visualization, virtual camera with picture overlay. Matching points widely spread in the X-direction of the image.}
	\label{FIG:sc005}
\end{figure}
\begin{figure}[htb]
	\centering
	\includegraphics[width=0.4\linewidth]{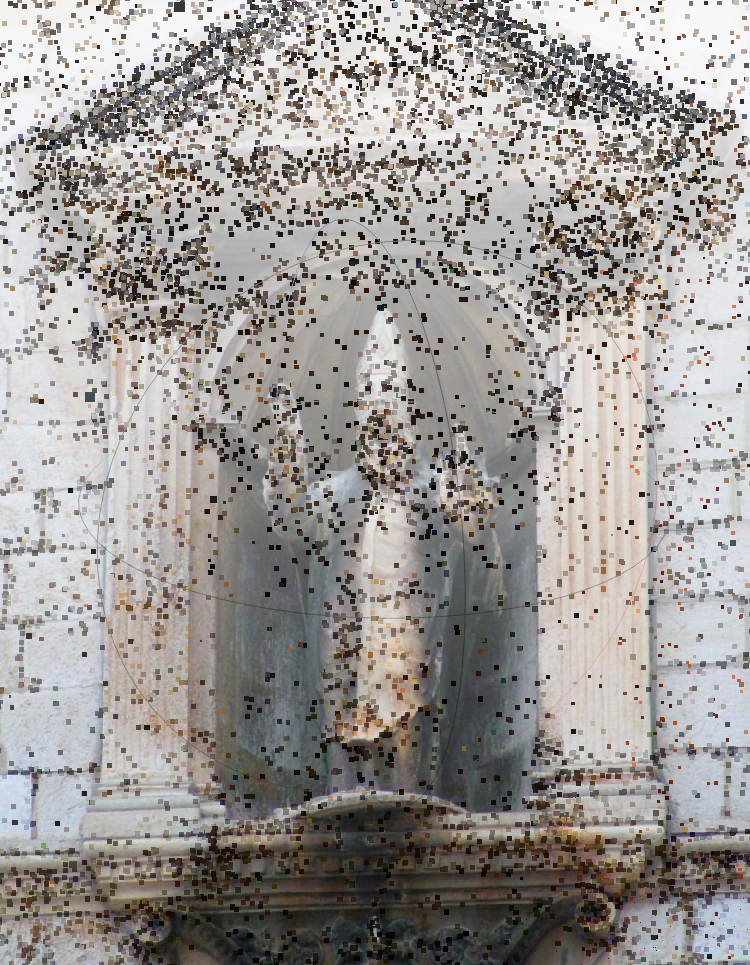}
	\caption{Meshlab visualization, virtual camera with picture overlay. Close-up shot, matches spread across the whole image.}
	\label{FIG:sc006}
\end{figure}

Most of the pictures that are not posed correctly now have colors that are off, a lot of people or other covering objects in them or generally do not have enough matched features across the image.

\begin{figure}[H]
	\centering
	\includegraphics[width=0.7\linewidth]{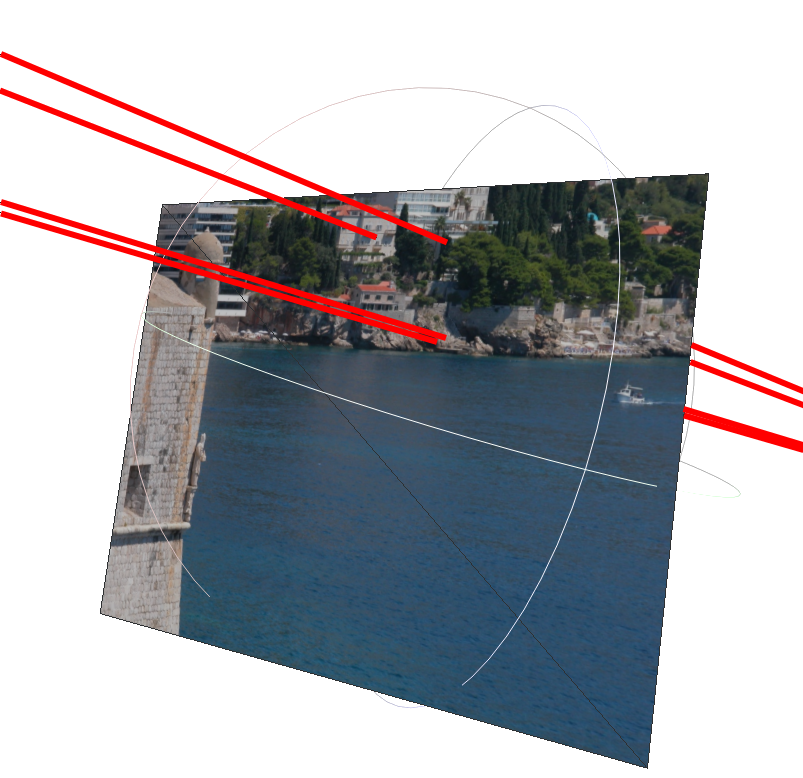}
	\caption{Example of a problematic query: Matches only in the far plane, not spread across the image due to a lot of water.}
	\label{FIG:sc010}
\end{figure}

\label{benchmark}
\section{Benchmarks}
\begin{figure}[H]
	\centering
	\includegraphics[width=0.8\linewidth]{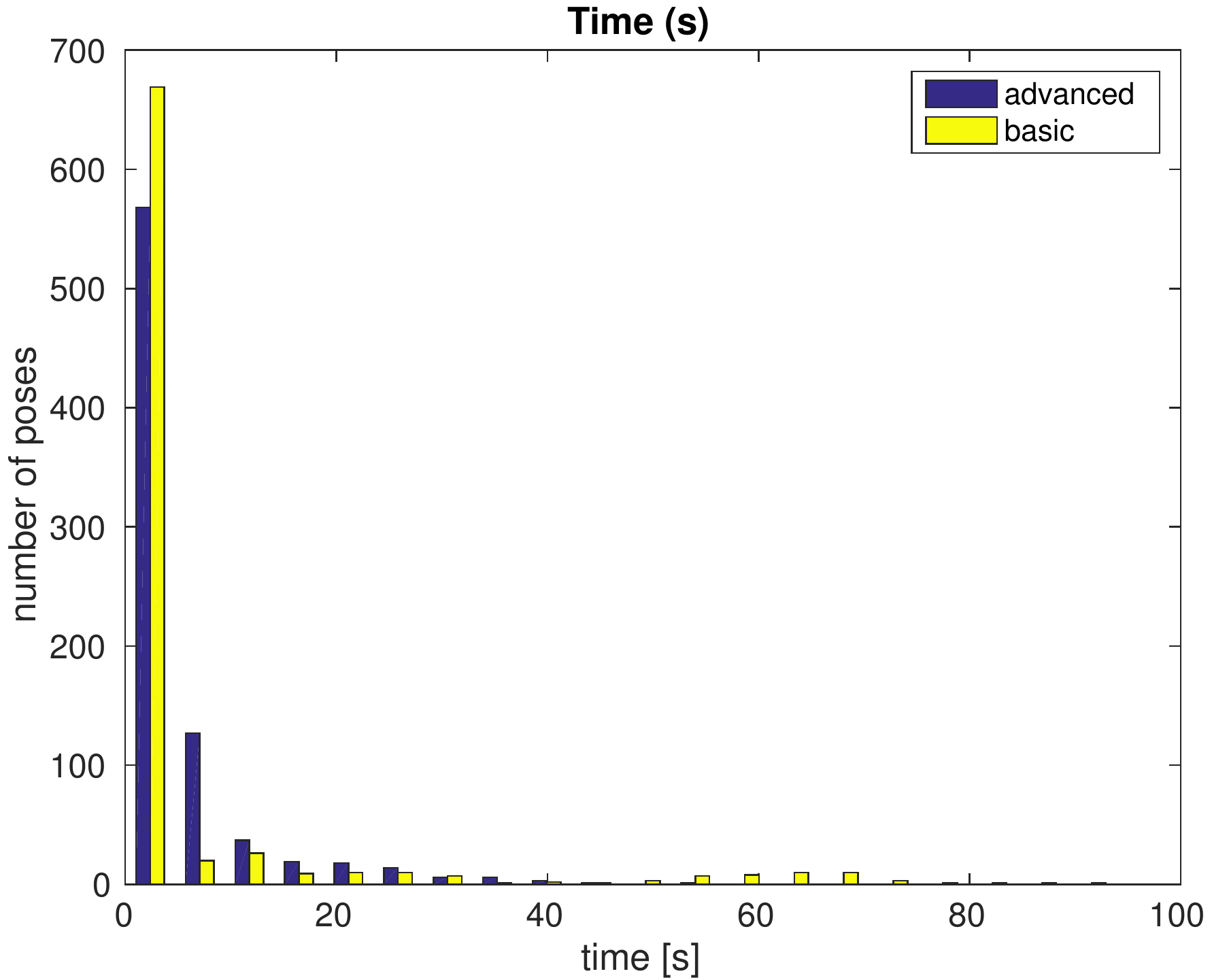}
	\caption{Benchmark timing results}
	\label{FIG:time}
\end{figure}

The benchmarks were carried out on an Intel i7 4790K processor and 16 GB RAM on Linux. We tuned both algorithms to a good trade-off between accuracy and running time. This resulted in similar timings: The basic method has on average $6.4\,s$ for a pose estimation, while the advanced method took $5.3\,s$. On problematic images, for which the basic method had to run up to $10'000$ RANSAC steps, the advance method uses exactly $200$ RANSAC steps, with a backmatching step after $100$ RANSAC steps. This results in longer runtime (up to $20\,s$), but is still better than many more RANSAC steps (up to $80\,s$).
It should be noted that the advanced approach takes a bit longer for good cases because it always runs through all RANSAC steps. This is traded against better accuracy. Running through all RANSAC steps on the basic method however would not be practical, as it would result in run times of about 60 seconds, as seen in the histogram for bad cases (figure \ref{FIG:time}).

\begin{figure}[H]
	\centering
	\includegraphics[width=0.8\linewidth]{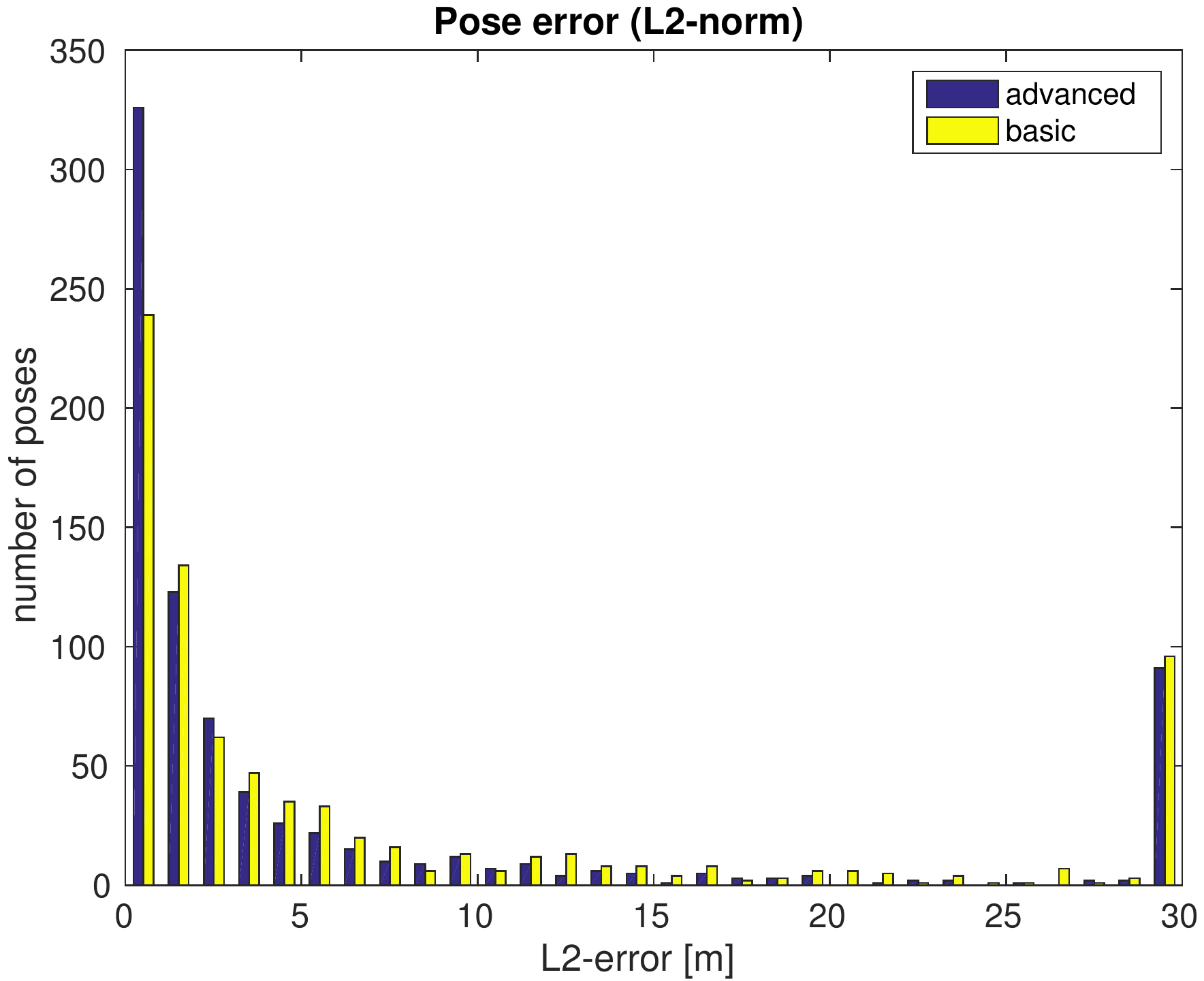}
	\caption{L2 pose error compared to golden poses}
	\label{FIG:pose_L2}
\end{figure}

In the $L_2$-error histogram (figure \ref{FIG:pose_L2}), we categorize poses with $\geq 30\,m$ error as completely wrong poses. The improvement on those hard cases was not significant from the basic to advanced method. However, backmatching still recovered some of the bad cases, but the effect was rather minor.

Running the RANSAC for a full 100 rounds before returning the best pose, combined with finding better points for pose estimation with co-occurence pushed a lot of already good poses to be more precise. The basic method has 16.75\% of poses with less than $0.5\,m$ error, the advanced 27.12\%. The median error on the basic method is  $2.42\,m$ and the mean $33.0\,m$, and on the advanced $1.6\,m$ median, $32.5\,m$ mean.

To understand better why some poses are completely wrong, looking at the given and estimated focal lengths gave additional insights. Of the 90 completely wrong poses (advanced), 34 also had focal lengths off more than 1000 pixels. Those are accounted both to wrong P4P estimation and difference from how bundler estimated the focal length compared to EXIF information given in the pictures.
With more than 1000 pixels off, the average error of poses was $174\,m$, while it was only $18\,m$ for those with less than 1000 pixels off focal lengths.

\begin{figure}[H]
	\centering
	\includegraphics[width=0.8\linewidth]{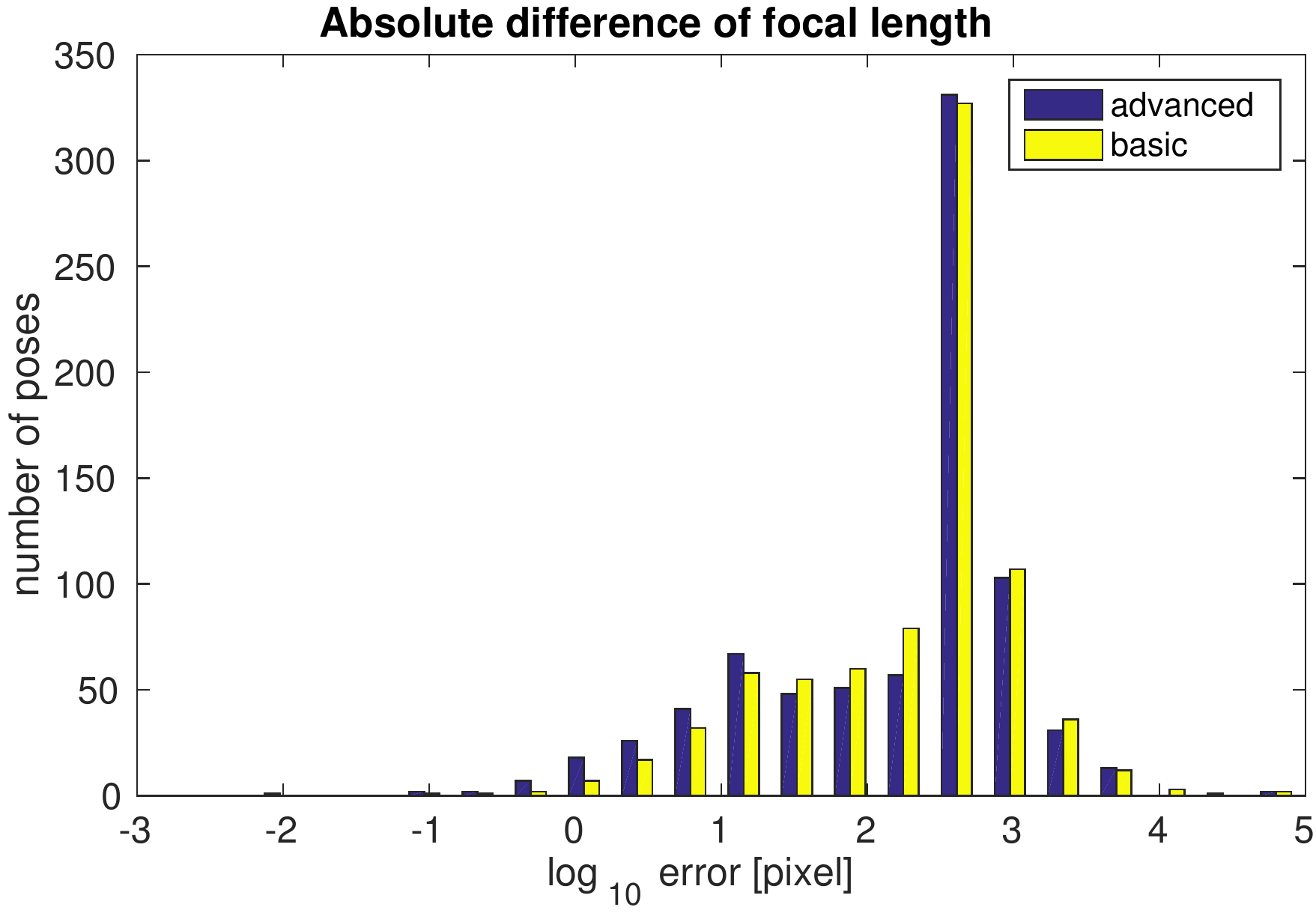}
	\caption{Focal length difference to golden pose}
	\label{FIG:focal}
\end{figure}

\section{Conclusions}
The original paper by Y. Li \etal \cite{li2010} claims to have 100\% registration rate on Dubrovnik and they compared their pose error on a different dataset (Quad) and against GPS. We, on the other hand, compare against the golden pose from the bundler output.

Both approaches should be taken with a grain of salt, as not all pictures have GPS and it is hard to match 3D point cloud coordinates to GPS coordinates. Also, the camera pose that bundler found in the Structure-from-Motion itself might be wrong.

Despite all of this, also our algorithm managed to find a pose for all query images, and most of them were somewhat correct (area, direction) even when being very off in terms of $L_2$ distance.
One number that closely coincided between the original paper and our approach was the median error of the poses: between $1.5\,m$ and $1.9\,m$ for them vs. $1.6\,m$ with our approach. We think this is indicative that we matched their quality, despite having different evaluations and datasets.

An outlook on possible improvements:
The poses could be further improved by two techniques:
\begin{enumerate}[nolistsep,noitemsep]
	\item Rerunning RANSAC on the fitted matches only, to stabilize towards a better solution in terms of points selected for pose estimation.
	\item Do final bundle adjustment to refine the precision locally within the best selected pose.
\end{enumerate}
However, also that would not help on completely wrong poses. Possibilities for that are:
\begin{enumerate}[nolistsep,noitemsep]
	\item Match against individual descriptors in a bigger FLANN kd-tree instead of averaged descriptors.
	\item Add additional priors to select points from the good matches for pose estimation.
	\item Make P3P and P4P more stable by adding scene understanding, such as assumptions about shot angle and image upside orientation.
	\item As an addition to the last point, try both P3P and P4P even in the case of known (possibly wrong) focal lengths and accept the one giving better estimates for the pose.
\end{enumerate}

\newpage

\section{Source Code}
We provide our software as open source:\\
\url{www.github.com/naibaf7/pose_estimation}

\section*{References}
\printbibliography[heading=none]

\end{document}